\documentclass{article}

\usepackage{microtype}
\usepackage{graphicx}
\usepackage{subcaption}
\usepackage{booktabs} 

\usepackage{hyperref}

\usepackage[preprint]{icml2026}

\usepackage{amsmath}
\usepackage{amssymb}
\usepackage{mathtools}
\usepackage{amsthm}

\usepackage[capitalize,noabbrev]{cleveref}

\usepackage{multirow}

\theoremstyle{plain}

\theoremstyle{definition}

\theoremstyle{remark}

\usepackage[textsize=tiny]{todonotes}

\icmltitlerunning{Empirical-MCTS: Continuous Agent Evolution via Dual-Experience Monte Carlo Tree Search}
\begin{document}
\twocolumn[
\icmltitle{Empirical-MCTS: Continuous Agent Evolution via Dual-Experience Monte Carlo Tree Search}
\begin{icmlauthorlist}
\icmlauthor{Hao Lu}{inst1}
\icmlauthor{Haoyuan Huang}{inst1}
\icmlauthor{Yulin Zhou}{inst1}
\icmlauthor{Chen Li}{inst1}
\icmlauthor{Ningxin Zhu}{inst1}
\end{icmlauthorlist}

\icmlaffiliation{inst1}{JianChengXingYun Technology Co., Ltd., Shenzhen, China}
\icmlcorrespondingauthor{Hao Lu}{luhao@jianchengxingyun.com}
\icmlcorrespondingauthor{Chen Li}{lichen@jianchengxingyun.com}

\icmlkeywords{Large Language Models, Multi-Model Systems, Adaptive Model Selection, Monte Carlo Tree Search, Preference Learning, Evolutionary Meta-Prompting}
\vskip 0.3in
]

\printAffiliationsAndNotice{}  

\begin{abstract}
Inference-time scaling strategies, particularly Monte Carlo Tree Search (MCTS), have significantly enhanced the reasoning capabilities of Large Language Models (LLMs). However, current approaches remain predominantly stateless, discarding successful reasoning patterns after each problem instance and failing to mimic the empirical accumulation of wisdom characteristic of human problem-solving. To bridge this gap, we introduce Empirical-MCTS, a dual-loop framework that transforms stateless search into a continuous, non-parametric learning process. The framework unifies local exploration with global memory optimization through two novel mechanisms: Pairwise-Experience-Evolutionary Meta-Prompting (PE-EMP) and a Memory Optimization Agent. PE-EMP functions as a reflexive optimizer within the local search, utilizing pairwise feedback to dynamically synthesize adaptive criteria and evolve meta-prompts (system prompts) in real-time. Simultaneously, the Memory Optimization Agent manages a global repository as a dynamic policy prior, employing atomic operations to distill high-quality insights across problems. Extensive evaluations on complex reasoning benchmarks, including AIME25, ARC-AGI-2, and MathArena Apex, demonstrate that Empirical-MCTS significantly outperforms both stateless MCTS strategies and standalone experience-driven agents. These results underscore the critical necessity of coupling structured search with empirical accumulation for mastering complex, open-ended reasoning tasks.
\end{abstract}

\section{Introduction}
\label{sec:introduction}

Large Language Models (LLMs) have demonstrated that increasing computational capability at inference time can significantly improve reasoning performance \cite{openai2024openaio1card,openai2025competitiveprogramminglargereasoning}. Techniques such as "Best-of-N" sampling \cite{brown2024largelanguagemonkeysscaling,Li_2022,lightman2023letsverifystepstep} and Monte Carlo Tree Search (MCTS) \cite{zhou2024languageagenttreesearch,feng2024alphazeroliketreesearchguidelarge,xu2023traingainunleashmathematical,zhang2024accessinggpt4levelmathematical,zhang2024llamaberrypairwiseoptimizationo1like,inoue2025wider,lu2025mctsrzeroselfreflectivepsychologicalcounseling,xie2024montecarlotreesearch} allow models to explore multiple reasoning paths and self-correct, effectively scaling performance without the need for expensive parameter updates. However, a critical limitation of current inference-time scaling methods is that they are stateless. Whether using standard MCTS or adaptive branching strategies \cite{inoue2025wider}, the agent treats each new problem as an isolated event. Once a search process concludes, the successful strategies or valid reasoning patterns discovered are discarded. This contrasts with human problem-solving, which is empirical: experts solve problems by combining long-term experience (accumulated domain knowledge) with short-term experience (immediate feedback and context from the current problem). 

Existing methods attempting to address this lack of memory often fall short in integration. Systems like FLEX \cite{cai2025flexcontinuousagentevolution} maintain an experience library but treat retrieval and reasoning as separate steps, preventing the model from evolving its search strategy dynamically. Other approaches, such as Training-Free GRPO \cite{cai2025trainingfreegrouprelativepolicy}, use historical data to adjust the generation probability but lack the structured exploration provided by tree search, limiting their effectiveness on complex, multi-step logic tasks. 

To address these limitations, we propose Empirical-MCTS, a framework that integrates continuous experience accumulation with structured tree search. Our approach unifies two distinct types of experience via a dual-loop evolutionary mechanism:

Short-term Experience (PE-EMP): Within the local search process, we introduce Pairwise-Experience-Evolutionary Meta-Prompting (PE-EMP). Instead of using static prompts for node expansion, PE-EMP functions as a reflexive optimizer. It analyzes pairwise response differences to synthesize adaptive criteria and dynamically evolves the meta-prompt (system prompt). This ensures that immediate feedback is not merely used for selection, but actively refines the generation policy for subsequent steps \cite{liu2025inferencetimescalinggeneralistreward}.

Long-term Experience (Memory Optimization): To sustain learning across problems, we implement a Memory Optimization Agent. Rather than a static retrieval database, our repository is treated as a dynamic policy prior. We employ an optimizer-based update mechanism—utilizing atomic operations such as add, modify, and merge—to continuously distill high-quality insights from the search process into a global knowledge base \cite{cai2025trainingfreegrouprelativepolicy}.

Furthermore, to ensure rigorous evaluation of the evolved prompts, we integrate a hybrid preference model adapted from recent pairwise optimization strategies \cite{zhang2024llamaberrypairwiseoptimizationo1like}. By mapping the explicit "self-principled" scores generated by PE-EMP into a globally consistent ranking via Enhanced Borda Count, we address the non-transitivity inherent in LLM preferences. This integration allows Empirical-MCTS to accurately distinguish and prioritize peak performance trajectories, ensuring that the qualitative insights gained from memory optimization are effectively translated into quantitative search guidance.

We evaluate Empirical-MCTS on diverse benchmarks, including AIME25 \cite{aime25}, ARC-AGI-2 \cite{chollet2026arcagi2newchallengefrontier}, and MathArena Apex \cite{balunovic2025matharena}, using DeepSeek-V3.1-Terminus \cite{deepseekai2024deepseekv3technicalreport}, gpt-oss-120b \cite{openai2025gptoss120bgptoss20bmodel} and Gemini 3 Pro \cite{google2025gemini3pro}. Empirical-MCTS achieved better results than previous approaches, such as repeated sampling, LLaMA-Berry, FLEX and Training-free GRPO. Our key contributions are:
\begin{itemize}
    \item We introduce Empirical-MCTS, a novel paradigm that bridges the gap between structured search and continuous learning. By unifying local exploration with global memory optimization, we transform MCTS from a stateless inference technique into a non-parametric online learning agent capable of accumulating "wisdom" across disparate problem instances without weight updates.
    \item We propose Pairwise-Experience-Evolutionary Meta-Prompting (PE-EMP). Moving beyond static expansion, PE-EMP acts as a reflexive optimizer that synthesizes "self-principles" from pairwise feedback. This mechanism dynamically evolves the meta-prompt (system prompt) in real-time, allowing the agent to actively refine its reasoning policy and navigate complex logic spaces with increasing precision during the search.
    \item We show that Empirical-MCTS outperforms standard MCTS and previous experience-based models. It achieves state-of-the-art results on high-difficulty benchmarks like AIME25, ARC-AGI-2, and MathArena Apex. Our results prove that "remembering" past reasoning patterns is the key to solving complex, multi-step logic problems more efficiently than traditional search methods.
\end{itemize}

\section{Related Work}
\label{sec:related_work}

\subsection{Inference-Time Scaling and Tree Search}
Inference-time scaling focuses on improving model outputs by allocating more compute during generation. Early methods, such as Self-Consistency \cite{wang2023selfconsistencyimproveschainthought}, generate multiple independent answers and select the most frequent one. To handle more complex tasks, tree-search algorithms like Tree of Thoughts (ToT) \cite{NEURIPS2023_271db992} and MCTS applied to LLMs \cite{zhou2024languageagenttreesearch,feng2024alphazeroliketreesearchguidelarge,xu2023traingainunleashmathematical,zhang2024accessinggpt4levelmathematical,zhang2024llamaberrypairwiseoptimizationo1like,inoue2025wider,lu2025mctsrzeroselfreflectivepsychologicalcounseling,xie2024montecarlotreesearch} were developed to structure the reasoning process into steps. Recent work has focused on optimizing the search structure. AB-MCTS \cite{inoue2025wider} introduces adaptive branching and LLaMA-Berry \cite{zhang2024llamaberrypairwiseoptimizationo1like} integrates MCTS with a pairwise reward model to improve selection accuracy. However, despite these structural improvements, these methods remain stateless across problems. They do not retain the successful reasoning patterns discovered during the search. Empirical-MCTS builds upon the robust search backbone of LLaMA-Berry but introduces a stateful paradigm, persisting empirical experience to evolve the agent's policy continuously.

\subsection{Experience-Driven Agents}
Several studies aim to enable agents to learn from interaction without parameter updates. FLEX \cite{cai2025flexcontinuousagentevolution} proposes a "Forward Learning" paradigm that stores successful trajectories in a library for future retrieval. However, FLEX largely relies on static retrieval mechanisms and does not deeply integrate this experience into the step-by-step decision-making of a search tree. Training-Free GRPO \cite{cai2025trainingfreegrouprelativepolicy} takes a different approach by acting as an "inference-time optimizer," using the LLM to analyze group results and update the context effectively. While this avoids gradient updates, it operates primarily as a policy adjustment tool rather than a reasoning search engine. Empirical-MCTS bridges these approaches by using MCTS to generate high-quality reasoning data and a global repository to store and refine this data, ensuring the agent becomes more efficient over time.

\subsection{Meta-Reasoning and Refinement}
Iterative refinement techniques, such as Self-Refine \cite{NEURIPS2023_91edff07} and Reflexion \cite{NEURIPS2023_1b44b878}, allow models to critique their own outputs. However, standard self-refinement can sometimes lead to the model discarding correct intermediate steps or drifting away from the original problem constraints. MCTSR-Zero \cite{lu2025mctsrzeroselfreflectivepsychologicalcounseling} suggests using "Meta-Prompts"—high-level instructions generated by the model—to guide subsequent generation. We formalize this as our "Short-term Experience" module. In our MCTS expansion step, we use explicit meta-prompts containing critiques and suggestions to generate child nodes. This ensures that the reasoning process remains focused and builds constructively on previous steps.

\section{Methodology: The Empirical-MCTS Framework}
\label{sec:methodology}

\subsection{Pairwise-Experience-Evolutionary Meta-Prompting (PE-EMP)}
\label{subsec:pe_emp}

\begin{figure*}[h]
\centering
\includegraphics[width=0.8\linewidth]{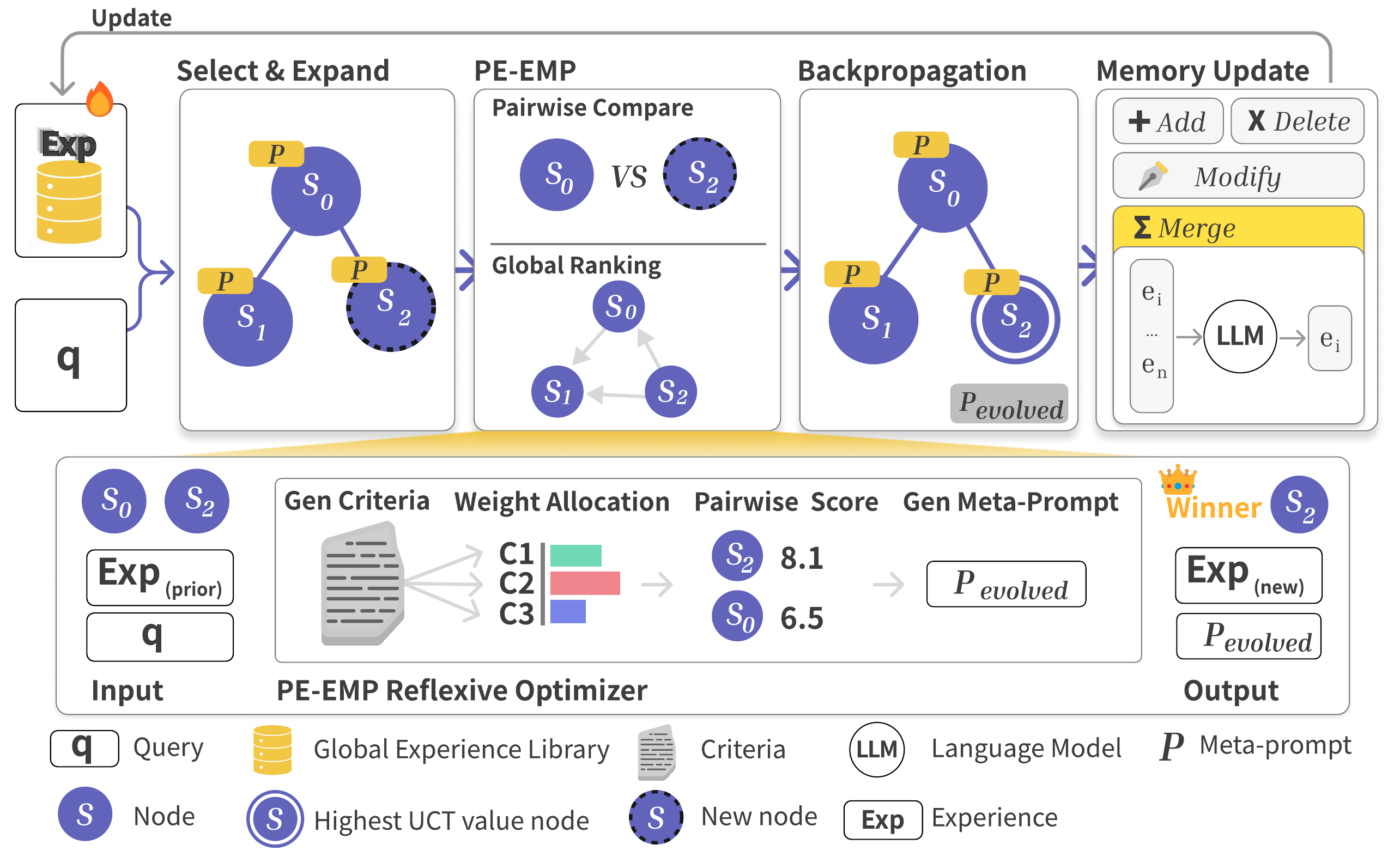}
\caption{Concrete Instantiation of Empirical-MCTS Framework.}
\label{fig:main_img_icml}
\end{figure*}

In Empirical-MCTS, the evolution of the meta-prompt (system prompt) is not an isolated heuristic but a derivative of the reward modeling process. Drawing inspiration from the Self-Principled Critique Tuning (SPCT) framework proposed for inference-time scaling of generalist reward models \cite{liu2025inferencetimescalinggeneralistreward}, we introduce a unified mechanism termed Pairwise-Experience-Evolutionary Meta-Prompting (PE-EMP).

SPCT demonstrates that generating specific "self-principles" prior to critiquing significantly enhances the scalability and accuracy of reward models. We extend this paradigm by transforming the Judge model from a mere discriminator into a reflexive optimizer. While SPCT utilizes principles to scale inference accuracy via sampling, PE-EMP leverages these generated principles (Adaptive Criteria) to drive the evolution of the generation policy itself.

Formally, let a node expansion result in a pairwise comparison between a candidate state $S_c = (\mathcal{P}_c, r_c)$ and a baseline (parent) state $S_p = (\mathcal{P}_p, r_p)$, where $\mathcal{P}$ denotes the meta-prompt and $r$ the model response. The PE-EMP module $\mathcal{J}$ processes the query $q$, prior accumulated experiences $\mathcal{E}_{prior}$, and the two states to yield a triadic output:

\begin{equation}
\mathcal{J}(q, \mathcal{E}_{prior}, S_c, S_p) \rightarrow (\mathbf{s}, \mathcal{E}_{new}, \mathcal{P}_{evolved})
\end{equation}

where $\mathbf{s}$ is the raw score vector, $\mathcal{E}_{new}$ denotes the new empirical insights derived from practice, and $\mathcal{P}_{evolved}$ is the evolved meta-prompt. The internal logic of the PE-EMP executes a strict seven-stage cognitive process that mirrors and extends the SPCT pipeline:
\begin{algorithm}[ht]
\caption{Empirical-MCTS: Continuous Agent Evolution}
\label{alg:empirical_mcts}
\begin{algorithmic}[1]
\STATE {\bfseries Input:} Query $q$, Initial Prompt $\mathcal{P}_0$, Global Experience Library $\mathcal{D}$, Iterations $T$
\STATE {\bfseries Initialize:} Search tree $\mathcal{T}$ with root $v_0(q, \mathcal{P}_0)$
\STATE $\mathcal{P}_{evolved} \leftarrow \mathcal{P}_0$ \COMMENT{Initialize evolved prompt with base prompt}
\STATE $\mathcal{E}_{prior} \leftarrow \text{Retrieve}(\mathcal{D}, q)$ \COMMENT{Retrieve long-term experience}
\FOR{$t = 1$ {\bfseries to} $T$}
    \STATE $S_p \leftarrow \text{Selection}(\mathcal{T}, \text{UCB})$ \COMMENT{Navigate tree nodes}
    
    \STATE \COMMENT{--- Local Loop: PE-EMP Expansion ---}
    \STATE $S_c \leftarrow \text{Expand}(S_p, \mathcal{P}_{evolved}, \mathcal{E}_{prior})$ \COMMENT{Generate child state}
    \STATE $(\mathbf{s}, \mathcal{E}_{new}, \mathcal{P}_{evolved}) \leftarrow \mathcal{J}(q, \mathcal{E}_{prior}, S_c, S_p)$ \COMMENT{PE-EMP Reflexive Optimizer}
    \STATE \textit{// PE-EMP internally: (1) Generate Criteria $\rightarrow$ (2) Critique $\rightarrow$ (3) Scoring $\rightarrow$ (4) Prompt Evolution}
    
    \STATE \COMMENT{--- Value Backpropagation ---}
    \STATE $R \leftarrow \text{Hybrid-Reward}(\mathbf{s}, \text{Borda-Count}(\mathcal{T}))$ \COMMENT{Section 3.3}
    \STATE $\text{Update-Q-Values}(\mathcal{T}, S_p, R)$ \COMMENT{Decay-based backprop}
    
    \STATE \COMMENT{--- Global Loop: Memory Update ---}
    \IF{$\mathcal{E}_{new}$ contains high-value insights}
        \STATE $\pi_{mem} \leftarrow \text{Optimizer}(\mathcal{D}, \mathcal{E}_{new})$ \COMMENT{Determine Add/Modify/Merge/Delete}
        \STATE $\mathcal{D} \leftarrow \text{Update-Library}(\mathcal{D}, \pi_{mem})$ \COMMENT{Non-parametric policy update}
    \ENDIF
\ENDFOR

\STATE {\bfseries Output:} Best response $r^*$ from $\mathcal{T}$
\end{algorithmic}
\end{algorithm}
\begin{enumerate}
    \item \textbf{Adaptive Criteria Generation (Self-Principled Phase)}: Aligning with SPCT's methodology, the model first acts as a principle generator. It dynamically formulates specific evaluation dimensions tailored to the user's query, ensuring that the subsequent critique is grounded in context-aware standards rather than generic rules.
    \item \textbf{Comparative Chain-of-Thought (Critique Phase)}: A step-by-step critical analysis is performed, contrasting the candidate and baseline responses against the generated criteria.
    \item \textbf{Dynamic Weight Allocation}: Weights are assigned to each criterion based on their relevance to the specific problem context, ensuring $\sum w_i = 100\%$.
    \item \textbf{Weighted Scoring}: A quantitative calculation determines the performance scores for both responses based on the weighted criteria.
    \item \textbf{Raw Score Generation}: The model performs a step-by-step arithmetic aggregation, calculating the weighted sum of dimension-specific scores derived in the previous step. This yields a precise scalar value within the range $[0, 10]$ for each response, serving as the high-fidelity signal for probabilistic preference modeling.
    \item \textbf{Context-Aware Insight Synthesis}: Building upon the \textit{prior experiences} retrieved from the RAG repository, the model synthesizes the immediate feedback with historical wisdom. It distills the current interaction into new, high-order empirical insights ($\mathcal{E}_{new}$), explicitly identifying how the current case refines, expands, or corrects the existing knowledge boundary provided by the retrieval system.
    \item \textbf{Strategic Prompt Evolution}: Finally, the model analyzes the fundamental execution differences between $\mathcal{P}_c$ and $\mathcal{P}_p$. By internalizing the \textit{Critical Success Factors} of the winner and the \textit{Failure Lessons} of the loser, it synthesizes a more advanced meta-prompt $\mathcal{P}_{evolved}$.
\end{enumerate}

This architecture ensures that $\mathcal{P}_{evolved}$ is not a random mutation but a conscious adaptation. By incorporating the \textit{self-principled} logic from SPCT, the meta-prompt explicitly encodes the high-quality evaluation criteria and lessons learned from the immediate pairwise battle, creating a directed evolutionary pressure toward prompts that produce higher-reward responses.

To prevent semantic drift in abstract reasoning tasks, we further filter retrieved experiences by task taxonomy alignment. Each experience in $\mathcal{D}$ is tagged with a coarse-grained task type (e.g., \texttt{geometry-proof}, \texttt{combinatorial-optimization}) derived from the original problem source. Only experiences sharing the same task type as $q$ (determined via zero-shot classification with the base LLM) are included in $\mathcal{E}_{prior}$. We set $k=5$ as the default retrieval size, balancing context richness against prompt length constraints.

\subsection{Experience Memory Optimization}
\label{subsec:experience_memory}

A critical contribution of Empirical-MCTS is its active management of the experience library, a methodology directly inspired by the Training-Free GRPO \cite{cai2025trainingfreegrouprelativepolicy}. Training-Free GRPO posits that LLM agents can achieve alignment effects analogous to parametric reinforcement learning (e.g., PPO \cite{schulman2017proximalpolicyoptimizationalgorithms}) by iteratively refining experiential knowledge as a "token prior" rather than updating model weights.

Adopting this paradigm, we implement a \textit{Memory Optimization Agent} that functions as the optimizer in the context space. Unlike static RAG systems that simply append inputs, this agent treats the distilled insights $\mathcal{E}_{new}$ as gradients for updating the global knowledge state.

Let $\mathcal{D}_t$ be the current experience library. Upon receiving the batch of new insights from the PE-EMP phase, the system invokes the optimization routine. Following the optimization protocol defined in Training-Free GRPO, the agent analyzes the new insights against the existing library $\mathcal{E}_{exist}$ and generates a structured update plan $\pi_{mem}$ consisting of four atomic operations:

\begin{itemize}
    \item \textbf{Add}: Append distinct, high-quality experiences described in the new insights to the library $\mathcal{D}_t$, effectively expanding the policy's action space.
    \item \textbf{Modify}: Refine or correct an existing experience $e_i \in \mathcal{D}_t$ based on the fresh "semantic advantage" derived from the current iteration, sharpening the policy's precision.
    \item \textbf{Merge}: Synthesize multiple fragmented experiences $\{e_i, e_j, ...\}$ into a single, cohesive principle to reduce context redundancy and improve retrieval efficiency.
    \item \textbf{Delete}: Prune obsolete or low-quality experiences that no longer align with high-reward outcomes, functioning as a "forgetting" mechanism for bad policies.
\end{itemize}

Formally, this step represents a non-parametric policy update:
\begin{equation}
\mathcal{D}_{t+1} \leftarrow \text{Optimizer}(\mathcal{D}_t, \pi_{mem}(\mathcal{E}_{new}, \mathcal{E}_{exist}))
\end{equation}

By continuously optimizing $\mathcal{D}$, Empirical-MCTS shifts the output distribution $\pi(y|q, \mathcal{D})$ towards higher rewards over time, achieving continuous self-improvement without the computational cost of gradient-based training.

\subsection{Value Estimation via Hybrid Preference Integration}
\label{subsec:preference_modeling}

To transform the discrete insights from PE-EMP into a continuous reward signal, we adapt the robust Pairwise Preference Reward Model (PPRM) architecture from LLaMA-Berry \cite{zhang2024llamaberrypairwiseoptimizationo1like}. Unlike previous approaches that rely on opaque logits from a separate reward model, our framework uniquely leverages the explicit "Self-Principled Scores" generated by PE-EMP (Section \ref{subsec:pe_emp}) as the high-fidelity input source. This integrates the fine-grained, instruction-following assessment of our reflexive optimizer with the global consistency guarantees of graph-theoretic ranking.

Local and Global Valuation. We map the raw score vector $\mathbf{s} = (S_c, S_p)$ from PE-EMP to a transition probability via the Bradley-Terry model: $Q_{local}(s_c) = \frac{\exp(S_c)}{\exp(S_c) + \exp(S_p)}$. To address the non-transitivity inherent in pairwise LLM preferences, we employ the Enhanced Borda Count (EBC) method \cite{zhang2024llamaberrypairwiseoptimizationo1like} to construct a global ranking from the pairwise transition graph. The final reward $R(s)$ is a weighted fusion of the local probability and the global Borda rank, ensuring that the evolved prompts are evaluated within the context of the entire search history.

\subsection{Search Strategy}
\label{subsec:bg_search_strategy}

We employ a standard Upper Confidence Bound (UCB) applied to Trees for node selection. For value backpropagation, we adopt the decay-based update rule from Self-Refine MCTS \cite{zhang2024llamaberrypairwiseoptimizationo1like}: $Q(s_p) \leftarrow (1-\gamma)Q(s_p) + \gamma Q(s_c)$. This "soft-max" mechanism is particularly essential for our framework, as it allows the sharp, high-reward insights discovered by the Memory Optimization Agent to propagate efficiently up the tree without being diluted by earlier, less experienced exploration attempts.

\section{Experiments}
\label{sec:experiments}
\subsection{Experimental Setup}
\label{subsec:experimental_setup}
\textbf{Benchmarks.} We evaluated Empirical-MCTS on three high-difficulty benchmarks designed to test frontier reasoning and generalization capabilities: AIME25 \cite{aime25}, MathArena Apex \cite{balunovic2025matharena}, and ARC-AGI-2 \cite{chollet2026arcagi2newchallengefrontier}. AIME25 represents a standard high-school competition mathematics benchmark. MathArena Apex is a dynamic evaluation framework specifically curated to mitigate data contamination—a pervasive issue in current LLM evaluation—by utilizing problems from recent competitions (e.g., CMIMC 2025, IMO 2025). It is currently the only benchmark evaluating proof-writing capabilities. ARC-AGI (Abstraction and Reasoning Corpus) measures general fluid intelligence, requiring agents to synthesize transformation programs from minimal examples, a task where traditional LLMs historically struggle.

\textbf{Models.} Experiments were conducted using a diverse set of frontier models to demonstrate model-agnostic applicability: DeepSeek-V3.1-Terminus \cite{deepseekai2024deepseekv3technicalreport}, gpt-oss-120b \cite{openai2025gptoss120bgptoss20bmodel}, and the Gemini 3 Flash \cite{google2025gemini3flash}.

\textbf{Baselines.} We benchmark Empirical-MCTS against a spectrum of inference-time strategies: (1) LLM with In-Context Learning (ICL) \cite{dong2024surveyincontextlearning}, (2) LLM agent with reasoning-and-acting workflow (ReAct) \cite{yao2023reactsynergizingreasoningacting}, (3) Repeated Sampling (Best-of-$N$) \cite{brown2024largelanguagemonkeysscaling,Li_2022,lightman2023letsverifystepstep}, (4) FLEX \cite{cai2025flexcontinuousagentevolution}, (5) LLaMA-Berry \cite{zhang2024llamaberrypairwiseoptimizationo1like}, and (6) Training-Free GRPO \cite{cai2025trainingfreegrouprelativepolicy}.

\subsection{Results on Mathematical Reasoning}
We first assess the impact of adding empirical memory to structured search on mathematical reasoning tasks. Table~\ref{tab:math_results} summarizes the results on AIME25 and MathArena Apex.

\textbf{Surpassing Stateless Search.} On AIME25, Empirical-MCTS achieves 73.3\%, outperforming both the stateless LLaMA-Berry framework (63.3\%) and the memory-augmented FLEX agent (66.6\%). This result validates our core hypothesis: By integrating memory optimization directly into the node expansion via PE-EMP, Empirical-MCTS allows the agent to navigate the solution space more efficiently.

\textbf{Solving the Unsolvable.} The results on MathArena Apex are noteworthy. The base model, DeepSeek-V3.1-Terminus, scored 0.00\% across 16 runs, indicating that the model lacks the requisite reasoning priors for these problems. While Repeated Sampling also failed (0.00\%) and LLaMA-Berry achieved only marginal success (2.08\%), Empirical-MCTS reached 4.17\%. While the absolute score is low due to the extreme difficulty of the benchmark, the relative improvement is substantial. It demonstrates that our framework does not merely extract existing knowledge but synthesizes new solution paths through the accumulation of empirical wisdom during the search.

\begin{table}[t]
\caption{Performance comparison on AIME25 and MathArena-Apex. \textbf{Baseline} denotes DeepSeek-V3.1-Terminus. All evaluated frameworks utilize this model as the underlying backbone.}
\label{tab:math_results}
\centering
\begin{small}
\begin{sc}
\setlength{\tabcolsep}{4pt} 
\begin{tabular}{lccc}
\toprule
Method / Framework & AIME25 & MathArena-Apex\\
\midrule
Baseline & 56.7 & 0.00 (16 runs) \\
ICL                    & 53.3 & -              \\
ReAct                  & 60.0 & -              \\
FLEX                   & 66.6 & -              \\
Repeated Sampling      & 70.0 & 0.00 (4 runs)  \\
LLaMA-Berry            & 63.3 & 2.08 (4 runs)  \\
\textbf{Ours}          & \textbf{73.3} & \textbf{4.17 (4 runs)} \\
\bottomrule
\end{tabular}
\end{sc}
\end{small}
\end{table}

\subsection{Efficiency and Cost Analysis}
A critical barrier to inference-time scaling is cost. In Table~\ref{tab:model_performance}, we compare Empirical-MCTS using the cost-efficient Gemini 3 Flash against state-of-the-art frontier models on MathArena Apex and ARC-AGI-2.

\textbf{High Performance at Low Cost.} Empirical-MCTS enables smaller, faster models to punch above their weight class. On MathArena Apex, Ours (Gemini 3 Flash) achieves an accuracy of 35.42\% with a total cost of only \$5.24. This outperforms Gemini 3 Pro, which scores 23.44\% at a cost of \$3.40, and drastically outperforms GPT-5.2 (High), which scores 13.54\% at a cost of \$12.00. Similarly, on ARC-AGI-2, our method achieves 38.33\%, surpassing the much more expensive Gemini 3 Pro (31.1\%) and Grok 4 (16.0\%). 

\begin{figure}[h]
\centering
\includegraphics[width=0.8\linewidth]{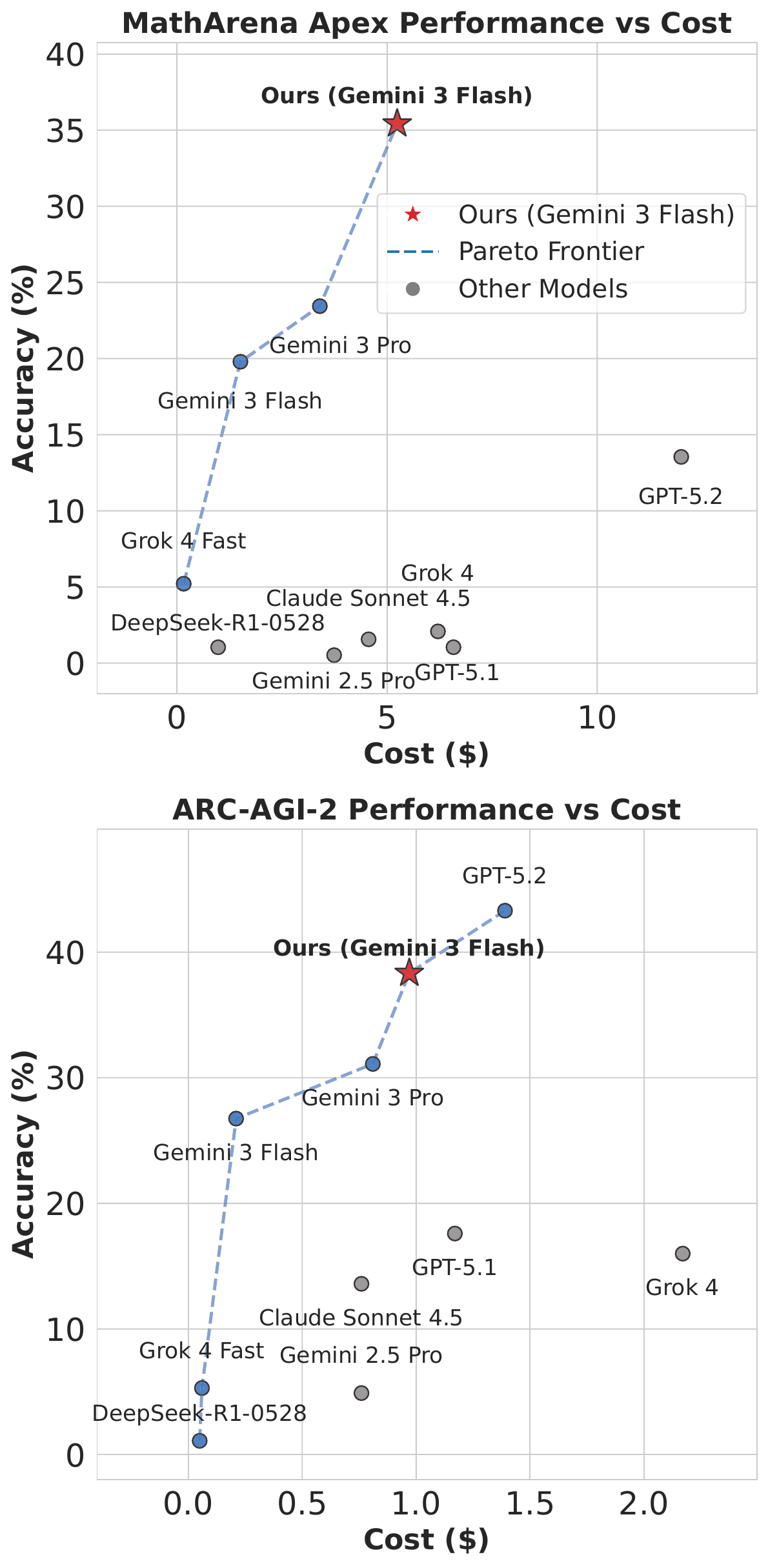}
\caption{\textbf{Cost-Performance Pareto Frontier Analysis.} We plot accuracy against inference cost for various models on MathArena Apex and ARC-AGI-2. The dashed blue line indicates the Pareto frontier, representing the optimal trade-off between cost and performance.}
\label{fig:pareto_frontier_analysis}
\end{figure}

Figure~\ref{fig:pareto_frontier_analysis} visualizes the Pareto frontier of reasoning efficiency. As shown in the upper panel (MathArena Apex), our approach strictly dominates all baselines, achieving the highest accuracy while maintaining a cost significantly lower than the heavy-compute models (e.g., GPT-5.2 at \$12.00). In the ARC-AGI-2 landscape, although GPT-5.2 holds a marginal accuracy lead, Empirical-MCTS resides on the "knee" of the curve—delivering comparable frontier-level reasoning at a reduced price point.

\begin{table*}[t]
\caption{Performance and cost analysis on MathArena Apex and ARC-AGI-2. Empirical-MCTS with Gemini 3 Flash outperforms more expensive models (e.g., GPT-5.2, Grok 4), establishing a new Pareto frontier for cost-effective reasoning.}
\label{tab:model_performance}
\centering
\begin{small}
\begin{sc}
\setlength{\tabcolsep}{8pt} 
\begin{tabular}{lcccc}
\toprule
\multirow{2}{*}{Model} & \multicolumn{2}{c}{MathArena Apex} & \multicolumn{2}{c}{ARC-AGI-2} \\
\cmidrule(lr){2-3} \cmidrule(lr){4-5}
 & Acc & Cost (\$) & Acc & Cost (\$) \\
\midrule
\textbf{Ours (Gemini 3 Flash)}  & \textbf{35.42\%} & 5.24 & 38.33\% & 0.97 \\
GPT-5.2 (High) \cite{openai2025gpt5_2}                  & 13.54\%          & 12.00        & \textbf{43.3}\%           & 1.39           \\
Gemini 3 Pro \cite{google2025gemini3pro}                    & 23.44\%          & 3.40         & 31.1\%           & 0.81          \\
Gemini 3 Flash \cite{google2025gemini3flash}                  & 19.79\%          & 1.51         & 26.75\%          & 0.21         \\
Grok 4 \cite{grok4}                         & 2.08\%           & 6.21         & 16.0\%           & 2.17           \\
GPT-5.1 (High) \cite{openai2025gpt5_1}                 & 1.04\%           & 6.58         & 17.6\%           & 1.17           \\
Claude Sonnet 4.5 \cite{claude_sonnet4_5}              & 1.56\%           & 4.56         & 13.6\%           & 0.76          \\
Grok 4 Fast (Reasoning) \cite{grok4fast}        & 5.21\%           & 0.16         & 5.3\%            & 0.06          \\
Gemini 2.5 Pro \cite{Gemini2_5pro}                 & 0.52\%           & 3.74         & 4.9\%            & 0.76          \\
DeepSeek-R1-0528 \cite{Guo_2025}               & 1.04\%           & 0.98         & 1.1\%            & 0.05          \\
\bottomrule
\end{tabular}
\end{sc}
\end{small}
\end{table*}

\subsection{Ablation Studies and Scalability Analysis}
\label{subsec:ablation}
To disentangle the contributions of PE-EMP, Memory Optimization, and Meta-Prompting, we conducted extensive ablation studies using gpt-oss-120b on the AIME25 benchmark. We also analyzed the scalability of our method with respect to the number of search rollouts.

\begin{figure}[h]
\centering
\includegraphics[width=0.9\linewidth]{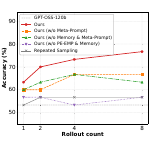}
\caption{Ablation study on AIME25 using gpt-oss-120b.}
\label{fig:ablation_curve}
\end{figure}

\textbf{Component Contribution and Scaling Behavior.}
Figure~\ref{fig:ablation_curve} presents the performance trajectory across different rollout configurations. Key observations include:
\begin{itemize}
    \item \textbf{Full Framework:} Achieves a peak of \textbf{76.7\%} at 8 rollouts, demonstrating continuous improvement (63.3\% $\rightarrow$ 76.7\%) that suggests further gains with increased compute.
    \item \textbf{w/o Meta-Prompt:} Removing explicit meta-prompting (with static prompts) in PE-EMP but retaining memory drops performance to 66.7\% at 8 rollouts. This indicates that while memory provides the \textit{content}, meta-prompting is essential for \textit{instructing} the model on how to apply that content effectively.
    \item \textbf{w/o Memory \& Meta-Prompt:} This configuration, which excludes the experience library and utilizes static prompts, peaks at 66.7\% (rollout 3) but then regresses to 63.3\% at rollout 8, failing to accumulate "wisdom" from previous rollouts.
    \item \textbf{w/o PE-EMP \& Memory:} To rigorously isolate the efficacy of the self-principled and pairwise comparison mechanisms, this variant removes the PE-EMP module, effectively degenerating into a standard MCTS framework. In this setting, the reward model utilizes static pre-defined criteria and assesses node quality pointwise scoring of individual responses, rather than through relative pairwise comparison. The resulting performance drop (to 56.7\% at rollout 8) mirrors that of naive repeated sampling, underscoring that the core engine of our improvement is the \textit{evolution} of the prompt and experience driven by pairwise signals, not merely the tree structure.
    \item \textbf{Repeated Sampling:} Plateaus at 56.7\% after 2 rollouts, confirming the limitations of stateless approaches.
\end{itemize}

\begin{figure}[h]
\centering
\includegraphics[width=0.8\linewidth]{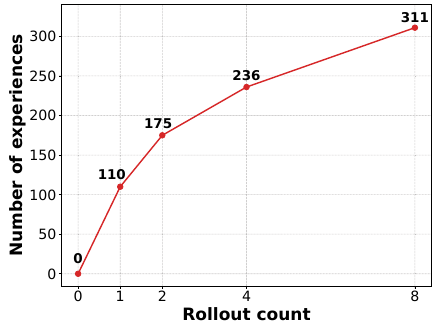}
\caption{\textbf{Growth of empirical memory repository during search.} The number of distilled experiences increases monotonically with rollout count (0 $\rightarrow$ 8), enabling progressive policy refinement across problem instances.}
\label{fig:experience_scaling}
\end{figure}

\textbf{Empirical Memory Growth Dynamics.}
Figure~\ref{fig:experience_scaling} quantifies how the global experience repository expands during search. Starting from zero experiences at initialization, the repository grows monotonically with each rollout (110 $\rightarrow$ 311 experiences after 8 rollouts), demonstrating the framework's capacity for continuous knowledge accumulation without parameter updates.

\begin{figure}[h]
\centering
\includegraphics[width=0.9\linewidth]{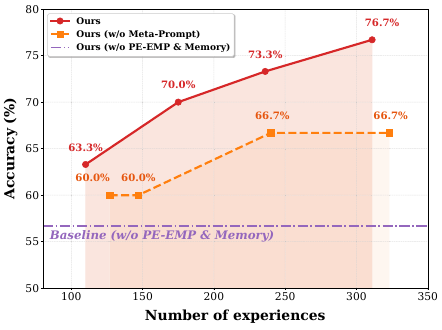}
\caption{\textbf{Performance vs. accumulated experiences on AIME25.} The full framework (red) demonstrates a strong positive correlation between experience count and accuracy, while ablated variants show limited improvement (yellow). Baseline performance shown as horizontal reference line (56.7\%).}
\label{fig:performance_vs_experience}
\end{figure}

\textbf{Performance-Experience Correlation.}
Figure~\ref{fig:performance_vs_experience} reveals a strong positive correlation between accumulated experiences and reasoning performance. The full framework (red) shows a clear upward trajectory (63.3\% $\rightarrow$ 76.7\%) as experiences grow from 110 to 311. In contrast, the variant without meta-prompting (yellow) exhibits diminishing returns after 240 experiences, while the baseline (w/o PE-EMP \& Memory at rollout 8) ultimately reached 56.7\%. This confirms that both memory accumulation and meta-prompting are essential for translating experiences into performance gains.

\subsection{Qualitative Analysis of Policy Evolution}
To understand \textit{how} Empirical-MCTS improves, we analyzed the evolution of meta-prompts in the MathArena Apex tasks. In initial rollouts, the prompt $\mathcal{P}_{evolved}$ focused on generic instructions (e.g., "Check your calculations"). By Rollout 4, after the Memory Optimization Agent had distilled failures from invalid proofs, the meta-prompt had evolved to include highly specific constraints: \textit{"For geometry proofs involving cyclic quadrilaterals, explicitly verify Ptolemy's inequality before assuming existence."} This transition from generic to domain-specific guidance, driven by the feedback loop between PE-EMP and the Memory Agent, explains the model's ability to solve problems that were previously intractable for the base model.

\section{Conclusion}
\label{sec:conclusion}

We presented Empirical-MCTS, a framework that unifies structured search with non-parametric online learning. By coupling Pairwise-Experience-Evolutionary Meta-Prompting (PE-EMP) with a dynamic global memory optimizer, our approach enables agents to accumulate empirical wisdom and refine their reasoning policies in real-time without gradient updates.

Empirical evaluations across AIME25, MathArena Apex, and ARC-AGI-2 confirm that this dual-loop mechanism establishes a new Pareto frontier for reasoning efficiency. Our results demonstrate that converting search history into actionable policy priors allows significantly smaller models to outperform larger, stateless baselines, validating that "remembering" reasoning patterns is as critical as the search capability itself.

\textbf{Limitations.} A primary limitation of our framework is the assumption that the accumulation of experience and the evolution of meta-prompts are inherently positive. In practice, this relies on the base model's capability to accurately verify reasoning steps. For extremely complex tasks that exceed the model's verification boundary, there is a risk of integrating low-quality insights or "hallucinated wisdom." This can lead to degenerative evolution, where the meta-prompt acts on false premises, steering the search towards suboptimal regions. Future research will focus on mitigating this risk by introducing robust consistency checks and uncertainty quantification into the memory optimization loop to filter toxic experiences. Additionally, we aim to extend Empirical-MCTS to multi-turn interactions and longer-horizon planning tasks, investigating how empirical memory can persist and adapt over extended operational contexts.

\bibliography{references}
\bibliographystyle{icml2026}

\end{document}